\documentclass[11pt,a4paper]{article}
\usepackage{amsfonts}
\usepackage{amsthm}
\usepackage{amsmath}
\usepackage{amssymb}
\usepackage{mathrsfs}
\usepackage{dsfont}   %%% fuer \N, \R etc
\usepackage{latexsym,epsf,graphicx}
\usepackage{caption}
\usepackage[colorlinks=false]{hyperref}
\usepackage[dvips]{color}
\usepackage{url}

\newtheorem{thm}{Theorem}

\newtheorem{defi}{Definition}

\def\openE{{{\rm I}\kern-.16em {\rm E}}}
\def\RR{\mathbb R}

\numberwithin{equation}{section}

\title{\textbf{Comparison theorems on large-margin learning}\footnotetext{\dag
		Corresponding author: Dao-Hong Xiang, Email: daohongxiang@zjnu.cn. \newline
		The work by J. Fan is partially supported by the Hong Kong RGC ECS grant
		22303518 and the NSF grant of China (No. 11801478). The work by
		D.~H. Xiang is supported by the National Natural Science Foundation of
		China under Grant 11871438
		and 11771120.}}

\author{\textbf{Jun Fan$^1$, Dao-Hong Xiang$^{2\,\dag}$}\\
	\normalsize{$^1$ Department of Mathematics, Hong Kong Baptist University, Kowloon, Hong Kong, China}\\
	\normalsize{$^2$ Department of Mathematics, Zhejiang Normal University,
		Jinhua, Zhejiang 321004, China} }
\date{}

\begin{document}

\maketitle

 \begin{abstract}
 	\noindent This paper studies binary classification problem associated with a family of loss functions called large-margin unified machines (LUM), which offers a natural bridge between distribution-based likelihood approaches and margin-based approaches. It also can overcome the so-called data piling issue of support vector machine in the high-dimension and low-sample size setting. In this paper we establish some new comparison theorems for all LUM loss functions which
 	play a key role in the further error analysis of large-margin learning algorithms.
 \end{abstract}


%%%%%%%%%%%%%%%%%%%%%%Introduction%%%%%%%%%%%%%
\section{Introduction}
Classification is a very important research topic in statistical machine learning. There are a large amount of literature on various classification methods, ranging from the very classical distribution-based likelihood approaches such as Fisher linear discriminant analysis (LDA) and logistic regression \cite{HastieTibshiraniFriedman2001}, to the margin-based approaches such as the well-known support vector machine (SVM) \cite{BoserGuyonVapnik1992, CortesVapnik1995}. Each type of classifiers has their own merits. Recently, Liu and his coauthors proposed in \cite{LiuZhangWu2011} the so-called large-margin unified machines (LUMs) which establish a unique transition between these two types of classifiers. As noted in \cite{MarronToddAhn2007}, SVM may suffer from data piling problems in the high-dimension low-sample size (HDLSS) settings, that is, the support vectors will pile up on top of each other at the margin boundaries when projected onto the normal vector of the separating hyperplane. Data piling is usually not desired for a classifier since it indicates that noise in the data can have an undue influence on the position of the separating hyperplane, which leads to a suboptimal performance of the classifier. The proposed distance-weighted discrimination (DWD) in \cite{MarronToddAhn2007} can overcome the data piling problem. Actually, both of SVM and DWD are special cases of LUMs. To the best our knowledge, a systematic error analysis is still lacking for LUMs from the  learning theory perspective.

In this paper, we focus on the binary classification problem. Let the input space $X$ be a compact domain of $\RR^d$ and the output space $Y=\{-1, 1\}$ representing the two classes. The model we take for sampling and measuring errors is based on a probability measure $P$ on $Z:=X\times Y.$
The learning target in binary classification is to find a classifier $\mathcal{C}: X\to Y$ such that for a new observation $(x, y)$ we have $\mathcal{C}(x)=y$ with high probability.
$$\mathcal{R}(\mathcal{C})=\hbox {Prob}\{\mathcal{C}(x)\neq y\}=\int_{X}P(y\neq \mathcal{C}(x)|x)dP_{X}$$ is called misclassification error which is used to
measure the performance of a classifier $\mathcal{C}.$ Here $P_{X}$ is the marginal distribution of $P$ on $X$ and $P(y|x)$ is the conditional distribution at $x\in X.$ The classifier minimizing the misclassification error is called the Bayes rule $f_c$ defined as
$$f_c(x)=\left\{\begin{array}{ll}
1, & \hbox{if   } P(y=1|x)\geq P(y=-1|x),\\
-1, & \hbox{otherwise}.
\end{array}\right.
$$

The classifiers considered here are induced by real-valued functions $f: X\to \RR$ as $\mathcal{C}_{f}=\hbox{sgn}(f)$ defined by $\hbox{sgn}(f)(x)=1$ if $f(x)\geq 0$ and $\hbox{sgn}(f)(x)=-1$ otherwise. In this paper, the real-valued functions are estimated by the LUMs based on a sample ${\bf z}=\{(x_i, y_i)\}_{i=1}^m\in Z^m$ drawn independently according to the probability measure $P.$ The LUM loss $V$ is defined as follows.
\begin{defi} For given $0\leq p\leq \infty$ and $0<q\leq \infty,$ the LUM loss $V: \mathbb{R}\rightarrow\mathbb{R}_+$ is defined by
	\begin{equation}\label{def.loss}
	V(t)=\left\{\begin{array}{ll}
	1-t, & \hbox{if } t<\frac{p}{1+p},\\
	\frac{1}{1+p}\Big(\frac{q}{(1+p)t-p+q}\Big)^q, & \hbox{if } t\geq \frac{p}{1+p}.
	\end{array}\right.
	\end{equation}
\end{defi}
The LUM loss is a family of convex loss functions, which not only covers some well-known loss functions such as hinge loss $V^{(h)}(t)=(1-t)_{+}$ for SVM with $p=\infty$ and $q>0,$ and the DWD loss \cite{MarronToddAhn2007, WangZou2018} with $p=q=1$:
$$V^{(D)}(t)=\left\{\begin{array}{ll}
1-t, & \hbox{if } t<\frac{1}{2},\\
\frac{1}{4t}, & \hbox{if } t\geq\frac{1}{2},
\end{array}\right.$$
it also includes many new losses such as the hybrid of SVM with AdaBoost with $q=\infty$ and $p\geq0$:
$$V^{(he)}(t)=\left\{\begin{array}{ll}
1-t, & \hbox{if } t<\frac{p}{1+p},\\
\frac{1}{1+p}\exp\{-[(1+p)t-p]\}, & \hbox{if } t\geq\frac{p}{1+p}.
\end{array}\right.$$
%\cite{Zhang2004,BartlettJordanMcauliffe2006,MarronToddAhn2007,WangZou2018}.
Denote the generalization error $\mathcal{E}(f)$ associated with the LUM loss functions by
\begin{equation}\label{def.samplefree}
{\cal E}(f)=\int_{Z}V(yf(x))\ dP(x,y)=\int_{X}\int_{Y}V(yf(x))dP(y|x)\ dP_{X}(x).
\end{equation}
Let $\eta(x)=P(y=1|x).$ The minimizer $f_P$ of ${\cal E}(f)$ over all measurable functions \cite{LiuZhangWu2011} for $0<q<\infty$ and $0\leq p<\infty$ is
\begin{equation}\label{def.target1}
f_P(x)=\left\{\begin{array}{ll}
-\frac{1}{1+p}\left[\Big(\frac{1-\eta(x)}{\eta(x)}\Big)^\frac{1}{q+1}q-q+p\right], & \hbox{if } 0\leq \eta(x)<\frac{1}{2},\\
%
%\big[-\frac{p}{1+p}, \frac{p}{1+p}\big], & \hbox{if } P(y=1|x)=\frac{1}{2},\\
%
\frac{1}{1+p}\left[\Big(\frac{\eta(x)}{1- \eta(x)}\Big)^\frac{1}{q+1}q-q+p\right], & \hbox{if } \frac{1}{2}\leq \eta(x)\leq 1.
\end{array}\right.
\end{equation}
For $q=\infty,$ the LUM loss reduces to the hybrid loss of the hinge loss and exponential loss with the minimizer below:
\begin{equation}\label{def.target3}
f_P(x)=\left\{\begin{array}{ll}
\frac{1}{1+p}\big[\ln\frac{\eta(x)}{1-\eta(x)}-p\big], & \hbox{if } 0\leq \eta(x)<\frac{1}{2},\\
\frac{1}{1+p}\big[\ln\frac{\eta(x)}{1-\eta(x)}+p\big]& \hbox{if } \frac{1}{2}\leq \eta(x)\leq 1.
\end{array}\right.
\end{equation}
For $p=\infty,$ the LUM loss reduces to the hinge loss for the SVM with the minimizer $f_c,$ i.e., the Bayes rule.

The performance of the classifier $\mathcal{C}_{f}$ induced by $f$ can be measured by the excess misclassification error $\mathcal{R}(\hbox{sgn}(f))-\mathcal{R}(f_c).$
It was first shown in \cite{Zhang2004} that for the hinge loss ($p=\infty$) and any measurable function $f: X\to \RR,$ $\mathcal{R}(\hbox{sgn}(f))-\mathcal{R}(f_c)\leq \mathcal{E}(f)-\mathcal{E}(f_c),$ which motivates us to achieve similar comparison theorems for the LUM loss.

%%%%%%%%%%%%%%%Comparison theorem%%%%%%%%%%%
\begin{thm}\label{thm.comparison1}
	Let $V$ be a LUM loss with $0<q\leq\infty$ and $0\leq p<\infty.$
	For any probability measure $P$ and any measurable function $f: X\to\RR,$

	(1) If $0<p<\infty$ and $0< q\leq\infty$, we have
	\begin{equation}\label{thm.com1}
	\mathcal{R}\big(\hbox{sgn}(f))-\mathcal{R}(f_c)\leq C_p\{\mathcal{E}(f)-\mathcal{E}(f_P)\},
	\end{equation}
	where $C_p=\frac{p+1}{p}.$
	
	(2) If $p=0$ and $0< q\leq\infty$, we have
	\begin{equation}\label{thm.com2}
	\mathcal{R}\big(\hbox{sgn}(f))-\mathcal{R}(f_c)\leq C_q\sqrt{\mathcal{E}(f)-\mathcal{E}(f_P)},
	\end{equation}
	where $C_q=2\sqrt{\frac{q+1}{q}}$ if $0<q<\infty$ and $C_q=\sqrt{2}$ if $q=\infty$.
\end{thm}

%%%%%%%%%%%%%%%%%%%%%%%%%%%%%%%
%%%%%%%%%%%%%%%%%%%%%%%%%%%%%%
%%%%%%%%%%%%%%%%%%%%%%%%%%%%%%
The above comparison theorem excludes the case of $p=\infty.$ As mentioned before, the LUM loss reduces to the hinge loss when $p=\infty.$ The comparison theorem for hinge loss has been firstly studies in \cite{Zhang2004}, which has the same form with (\ref{thm.com1}) except for the constant.

%%%%%%%%%%%%%%%%%%%%%
In Theorem \ref{thm.comparison1}, we notice that the comparison theorem for $p=0$ is worse than the one for $0<p\leq\infty.$ (\ref{thm.com2})  may be improved if the distribution measure $P$ satisfies a noise condition \cite{Tsybakov2004} below.
\begin{defi}
	Let $0\leq \tau\leq \infty.$ We say that probability measure $P$ satisfies a Tsybakov noise condition with exponent $\tau$ if there exists a constant $C_\tau$ such that
	\begin{equation}\label{def.noise}
	P_{X}(\{x\in X: |2\eta(x)-1|\leq C_{\tau}t\})\leq t^\tau, \forall t>0.
	\end{equation}
\end{defi}
The inequality \ref{def.noise} is always met when $\tau=0$. The other extreme is for $\tau=\infty$, the case when $\eta(x)$ stays away from $\frac12$ with probability one.
\begin{thm}\label{thm.comparison3}
	Let $V$ be a LUM loss with $p=0$ and $0<q\leq \infty.$
	For any probability measure $P$ satisfying (\ref{def.noise}) with $0<\tau<\infty$ and any measurable function $f: X\to\RR,$
	\begin{equation}\label{thm.com5}
	\mathcal{R}\big(\hbox{sgn}(f))-\mathcal{R}(f_c)\leq C_\tau^{-\frac{\tau}{\tau+2}}2^{1+\frac{(2q+1)(\tau+1)}{(q+1)(\tau+2)}}
	\Big(\frac{q+1}{q}\Big)^{\frac{\tau+1}{\tau+2}}(\mathcal{E}(f)-\mathcal{E}(f_P))^{\frac{\tau+1}{\tau+2}}.
	\end{equation}
\end{thm}

\section{Proof of Comparison Theorems}
\begin{proof}[{\bf Proof of Theorem \ref{thm.comparison1}}]
	Denote $X_c=\{x\in X: \hbox{sgn}(f)(x)\ne f_c(x)\}.$ The definition of the misclassification error tells us that
	\begin{equation*}
	\mathcal{R}\big(\hbox{sgn}(f))-\mathcal{R}(f_c)=\int_{X_c}(P(\hbox{sgn}(f)(x)\ne y|x)-P(f_c(x)\ne y|x)\big)\ dP_{X}(x).
	\end{equation*}	
	For $x\in X_c,$ it is easy to see that
	$P(\hbox{sgn}(f)(x)\ne y|x)=P(f_c(x)=y|x),$ which yields that
	\begin{equation*}
	P(\hbox{sgn}(f)(x)\ne y|x)-P(f_c(x)\ne y|x)=
	P(f_c(x)=y|x)-P(f_c(x)\ne y|x)=\big|2\eta(x)-1\big|.
	\end{equation*}
	As a result, we get that
	\begin{equation}\label{proof.mis}
	\mathcal{R}\big(\hbox{sgn}(f))-\mathcal{R}(f_c)=\int_{X_c}\big|2\eta(x)-1\big|\ dP_{X}(x).
	\end{equation}	
	
	Define $\Phi(t)=\eta(x)V(t)+(1-\eta(x))V(-t).$ Let $x\in X.$ If we denote
	\begin{equation}\label{proof.gene}
	\mathcal{E}(f|x):=\int_{Y}V(yf(x))\ dP(y|x)=\eta(x)V(f(x))+(1-\eta(x))V(-f(x)),
	\end{equation}
	then $\mathcal{E}(f|x)=\Phi(f(x)).$ It follows that $f_P$ is also the minimier of $\Phi.$
	
	We first consider the case when $0<q<\infty$. If $\eta(x)\geq\frac{1}{2},$ $R(x)=\Big(\frac{\eta(x)}{1- \eta(x)}\Big)^\frac{1}{q+1}\geq 1,$ which leads to $f_P(x)\geq \frac{p}{1+p}.$ The definitions of LUM loss $V(t)$ and $f_P$ yield
	\begin{eqnarray*}\label{proof.e1}
		\Phi(f_P(x))&=&\eta(x)\frac{1}{1+p}\Big(\frac{q}{(1+p)f_P(x)-p+q}\Big)^q+(1-\eta(x))(1+f_P(x))\\\nonumber
		&=& \eta(x)\frac{1}{1+p}R(x)^{-q}+(1-\eta(x))\big[1+\frac{1}{1+p}(R(x)q-q+p)\big]\\\nonumber
		&=&\frac{1+|2\eta(x)-1|}{2(1+p)}\Big(\frac{1-|2\eta(x)-1|}{1+|2\eta(x)-1|}\Big)^{\frac{q}{q+1}}\\\nonumber
		& &+\frac{1-|2\eta(x)-1|}{2}\Big[1+\frac{1}{1+p}\Big(\Big(\frac{1+|2\eta(x)-1|}{1-|2\eta(x)-1|}\Big)^{\frac{1}{q+1}}q-q+p\Big)\Big].
	\end{eqnarray*}
	
	If $\eta(x)< \frac{1}{2},$ $R(x)^{-1}=\Big(\frac{1-\eta(x)}{ \eta(x)}\Big)^\frac{1}{q+1}>1,$ which implies that $f_P(x)<-\frac{p}{1+p}.$ Therefore,
	\begin{eqnarray*}\label{proof.e2}
		\Phi(f_P(x))&=&\eta(x)(1-f_P(x))+(1-\eta(x))\frac{1}{1+p}\Big[\frac{q}{-(1+p)f_P(x)-p+q}\Big]^q\\\nonumber
		&=& \eta(x)\big[1+\frac{1}{1+p}(R(x)^{-1}q-q+p)\big]+(1-\eta(x))\frac{1}{1+p}R(x)^{q}\\\nonumber
		&=&\frac{1+|2\eta(x)-1|}{2(1+p)}\Big(\frac{1-|2\eta(x)-1|}{1+|2\eta(x)-1|}\Big)^{\frac{q}{q+1}}\\\nonumber
		& &+\frac{1-|2\eta(x)-1|}{2}\Big[1+\frac{1}{1+p}\Big(\Big(\frac{1+|2\eta(x)-1|}{1-|2\eta(x)-1|}\Big)^{\frac{1}{q+1}}q-q+p\Big)\Big].
	\end{eqnarray*}
	
	We are interested in getting a lower bound for $\Phi(0)-\Phi(f_P(x)).$ For simplicity, we denote $a=|2\eta(x)-1|$ and $g(a)=\Phi(0)-\Phi(f_P(x)), a\in [0, 1].$
	Taking derivative of $g$ with respect to $a$ yield that
	\begin{equation}\label{gfun}
	g^\prime(a)=\frac{1}{2}+\frac{p-q}{2(p+1)}+\frac{q}{2(p+1)}
	\Big(\frac{1+a}{1-a}\Big)^{\frac{1}{q+1}}
	-\frac{1}{2(p+1)}\Big(\frac{1-a}{1+a}\Big)^{\frac{1}{q+1}}.
	\end{equation}
	Case I. If $0<p<\infty,$ it is easy to find that
	\begin{equation*}
	g^\prime(a)\geq \frac{1}{2}+\frac{p-q}{2(p+1)}+\frac{q}{2(p+1)}-\frac{1}{2(p+1)}=\frac{p}{p+1}.\nonumber
	\end{equation*}
	The mean value theorem together with the definition of $f_P$ tells us that
	\begin{equation*}
	g(a)-g(0)\geq \frac{p}{p+1}a,
	\end{equation*}
	that is
	\begin{equation}\label{proof.generror1}
	\Phi(0)-\Phi(f_P(x))=1-\Phi(f_P(x))\geq \frac{p}{p+1}|2\eta(x)-1|
	\end{equation}
	
	Combining (\ref{proof.mis}) with (\ref{proof.generror1}), we find that
	\begin{eqnarray*}
		\mathcal{R}\big(\hbox{sgn}(f))-\mathcal{R}(f_c)&=&\int_{X_c}\big|2\eta(x)-1\big|\ dP_{X}(x)\\
		&\leq &\frac{p+1}{p}\int_{X_c}\{\Phi(0)-\Phi(f_P(x))\}\ dP_{X}(x).
	\end{eqnarray*}
	
	Case II. If $p=0,$ it follows from (\ref{gfun}) that
	$$g'(a)=\frac{1}{2}-\frac{q}{2}+\frac{q}{2}\Big(\frac{1+a}{1-a}\Big)^{\frac{1}{q+1}}
	-\frac{1}{2}\Big(\frac{1-a}{1+a}\Big)^{\frac{q}{q+1}}.$$
	Obviously,
	$$g'(a)\geq \frac{q}{2}\Big(\frac{1+a}{1-a}\Big)^{\frac{1}{q+1}}-\frac{q}{2}\geq \frac{q}{2}\Big[(1+a)^{\frac{1}{q+1}}-1\Big].$$
	The mean value theorem together with $0\leq a\leq 1$ yields that for $\theta\in (0, a)$
	$$(1+a)^{\frac{1}{q+1}}-1=\frac{1}{q+1}\Big(\frac{1}{1+\theta}\Big)^{\frac{q}{q+1}}a\geq \frac{1}{q+1}\Big(\frac{1}{2}\Big)^{\frac{q}{q+1}}a.$$
	On the basis of the above facts, we find that
	\begin{equation}\label{proof.comp3}
	\Phi(0)-\Phi(f_{P}(x))\geq \frac{q}{q+1}\Big(\frac{1}{2}\Big)^{\frac{2q+1}{q+1}}a^2=\frac{q}{q+1}
	\Big(\frac{1}{2}\Big)^{\frac{2q+1}{q+1}}|2\eta(x)-1|^2.
	\end{equation}
	
	Applying the Cauchy-Schwarz inequality and the fact that $P_X$ is a probability measure on $X,$ we get that
	\begin{eqnarray*}
		\mathcal{R}\big(\hbox{sgn}(f))-\mathcal{R}(f_c)&=&\int_{X_c}\big|2\eta(x)-1\big|\ dP_{X}(x)\\
		&\leq &\Big\{\int_{X_c}\big|2\eta(x)-1\big|^2\ dP_{X}(x)\Big\}^{1/2}\Big\{\int_{X_c}1\ dP_X(x)\Big\}^{1/2}\\
		&\leq & 2\sqrt{\frac{q+1}{q}}\Big\{\int_{X_c}(\Phi(0)-\Phi(f_{P}(x)))\ dP_X(x)\Big\}^{1/2}.
	\end{eqnarray*}

	The definition of $f_P$ implies that
	$f_P(x)<-\frac{p}{p+1}$ if $\eta(x)<\frac{1}{2}$ and
	$ f_P(x)\geq \frac{p}{p+1}$ if $\eta(x)\geq \frac{1}{2}.$
	Recalling the definition of $f_c,$ we find that $f_P$ has the same sign of $f_c,$ that is Fisher consistency in statistics. In addition, for any $x\in X_c,$ $f(x)f_c(x)\leq 0.$ Therefore, $f(x)f_P(x)\leq 0,$ which implies that
	$0$ is between $f(x)$ and $f_P(x).$ This fact together with the convexity of $\Phi$ yields that
	$\Phi(f_P(x))\leq \Phi(0)\leq \Phi(f(x)).$

	Therefore, we get that for $0<p<\infty$
	\begin{eqnarray*}
		\mathcal{R}\big(\hbox{sgn}(f))-\mathcal{R}(f_c)&\leq &\frac{p+1}{p}\int_{X_c}\{\Phi(0)-\Phi(f_P(x))\}\ dP_{X}(x)\\
		&\leq &\frac{p+1}{p}\int_{X_c}\{\Phi(f(x))-\Phi(f_P(x))\}\ dP_{X}(x)\\
		&\leq &\frac{p+1}{p} \{\mathcal{E}(f)-\mathcal{E}(f_P)\}.
	\end{eqnarray*}
	
	Similarly, the following holds true for the case of $p=0.$
	\begin{eqnarray*}
		\mathcal{R}\big(\hbox{sgn}(f))-\mathcal{R}(f_c)
		&\leq & 2\sqrt{\frac{q+1}{q}}\Big\{\int_{X_c}(\Phi(0)-\Phi(f_{P}(x)))\ dP_X(x)\Big\}^{1/2}\\
		&\leq & 2\sqrt{\frac{q+1}{q}}\Big\{\int_{X_c}(\Phi(f)-\Phi(f_{P}(x)))\ dP_X(x)\Big\}^{1/2}\\
		&=&  2\sqrt{\frac{q+1}{q}}\sqrt{\mathcal{E}(f)-\mathcal{E}(f_P)}.
	\end{eqnarray*}
	This completes part of the proof when $0\leq p<\infty$ and $0<q<\infty$.
	
	Now we consider the case when $q=\infty$. Notice that the target function for the case of $q\to\infty$ is
	\begin{equation*}
	f_P(x)=\left\{\begin{array}{ll}
	\frac{1}{1+p}\big[\ln\frac{\eta(x)}{1-\eta(x)}-p\big], & \hbox{if } 0\leq \eta(x)<\frac{1}{2},\\
	\frac{1}{1+p}\big[\ln\frac{\eta(x)}{1-\eta(x)}+p\big]& \hbox{if } \frac{1}{2}\leq \eta(x)\leq 1.
	\end{array}\right.
	\end{equation*}
	In the same fashion as before, we get that
	\begin{equation*}
	\Phi(f_P(x))=\frac{1-|2\eta(x)-1|}{2}\Big(2+\frac{1}{p+1}\ln\frac{1+|2\eta(x)-1|}{1-|2\eta(x)-1|}\Big).
	\end{equation*}
	Let $a=|2\eta(x)-1|\in [0,1]$ and $g(a)=\Phi(0)-\Phi(f_{P}(x)).$ Then
	$$g(a)=1-\frac{1-a}{2}\Big(2+\frac{1}{p+1}\ln\frac{1+a}{1-a}\Big).
	$$
	Taking the derivative of $g(a)$ with respect to $a,$ we get that
	$$g'(a)=1+\frac{1}{2(p+1)}\ln\frac{1+a}{1-a}-\frac{1}{p+1}\frac{1}{1+a}.$$

	Case I. If $0<p<\infty,$ it is easy to get the lower bound of $g'(a)$ stated below.
	$$g'(a)\geq 1-\frac{1}{p+1}=\frac{p}{p+1}>0.$$
	
	Applying the mean value theorem and the definition of $f_P,$ we have
	$$\Phi(0)-\phi(f_P(x))\geq \frac{p}{p+1}|2\eta(x)-1|.$$
	
	Hence
	\begin{eqnarray*}
		\mathcal{R}\big(\hbox{sgn}(f))-\mathcal{R}(f_c)&=&\int_{X_c}\big|2\eta(x)-1\big|\ dP_{X}(x)\\
		&\leq &\frac{p+1}{p}\int_{X_c}\{\Phi(0)-\Phi(f_P(x))\}\ dP_{X}(x).
	\end{eqnarray*}

	Case II. If $p=0,$ the above inequality is not valid. In this case,
	$$g'(a)=1+\frac{1}{2}\ln\frac{1+a}{1-a}-\frac{1}{1+a}\geq 1-\frac{1}{1+a}\geq \frac{1}{2}a.$$

	Therefore,
	\begin{equation}\label{proof.comp4}
	\Phi(0)-\Phi(f_{P}(x))\geq \frac{1}{2}a^2=\frac{1}{2}|2\eta(x)-1|^2.
	\end{equation}
	
	Applying the Cauchy-Schwarz inequality and the fact that $P_X$ is a probability measure on $X,$ we get
	\begin{eqnarray*}
		\mathcal{R}\big(\hbox{sgn}(f))-\mathcal{R}(f_c)&=&\int_{X_c}\big|2\eta(x)-1\big|\ dP_{X}(x)\\
		&\leq &\Big\{\int_{X_c}\big|2\eta(x)-1\big|^2\ dP_{X}(x)\Big\}^{1/2}\Big\{\int_{X_c}1\ dP_X(x)\Big\}^{1/2}\\
		&\leq & \sqrt{2}\Big\{\int_{X_c}(\Phi(0)-\Phi(f_{P}(x)))\ dP_X(x)\Big\}^{1/2}.
	\end{eqnarray*}
	
	The desired results come immediately from the same discussion for $0<q<\infty$ case.
\end{proof}
\begin{proof}[{\bf Proof of Theorem \ref{thm.comparison3}}]
	Denote $X_t^{-}=\{x\in X: |2\eta(x)-1|\leq C_{\tau}t\}$ and  $X_t^{+}=\{x\in X: |2\eta(x)-1|\geq C_{\tau}t\}.$ We partition the set $X_c$ into $X_t^{-}$ and $X_t^{+}$ with a constant $t>0$ to be determined. On $X_t^{+},$ it follows that $|2\eta(x)-1|\leq \frac{|2\eta(x)-1|^2}{C_{\tau}t}.$ Combining (\ref{def.noise}) with the facts (\ref{proof.comp3}) and (\ref{proof.comp4}), we have
	\begin{eqnarray*}
		& &\mathcal{R}\big(\hbox{sgn}(f))-\mathcal{R}(f_c)
		=\int_{X_c}\big|2\eta(x)-1\big|\ dP_{X}(x)\\
		&=& \int_{x\in X_c\cap X_t^-}\big|2\eta(x)-1\big|\ dP_{X}(x)+\int_{x\in X_c\cap X_t^+}\big|2\eta(x)-1\big|\ dP_{X}(x)\\
		&\leq & \int_{x\in X_c\cap X_t^-}C_\tau t\ dP_{X}(x)+\int_{x\in X_c\cap X_t^+}\frac{\big|2\eta(x)-1\big|^2}{C_\tau t}\ dP_{X}(x)\\
		&\leq &C_\tau t P_{X}(\{x\in X: |2\eta(x)-1|\leq C_{\tau}t\})+\frac{ (q+1)2^{\frac{2q+1}{q+1}}}{qC_\tau t}\int_{X_c}(\Phi(0)-\Phi(f_P(x)))\ dP_X(x)\\
		&\leq & C_\tau t^{\tau+1}+\frac{ (q+1)2^{\frac{2q+1}{q+1}}}{qC_\tau t}\int_{X_c} (\Phi(0)-\Phi(f_P(x)))\ dP_X(x).
	\end{eqnarray*}
	Plugging  $t=\Big(\frac{(q+1)2^{\frac{2q+1}{q+1}}}{qC_\tau^2}\int_{X_c}(\Phi(0)-\Phi(f_P(x)))\ dP_X(x)\Big)^{\frac{1}{\tau+2}}$ into the above inequality, we have
	
	$$\mathcal{R}\big(\hbox{sgn}(f))-\mathcal{R}(f_c)\leq C_\tau^{-\frac{\tau}{\tau+2}}2^{1+\frac{(2q+1)(\tau+1)}{(q+1)(\tau+2)}}\Big(\frac{q+1}{q}\Big)^{\frac{\tau+1}{\tau+2}}
	\Big(\int_{X_c} (\Phi(0)-\Phi(f_P(x)))\ dP_X(x)\Big)^{\frac{\tau+1}{\tau+2}}.$$
	
	The definition of $f_P$ implies that
	$f_P(x)<-\frac{p}{p+1}$ if $\eta(x)<\frac{1}{2}$ and
	$ f_P(x)\geq \frac{p}{p+1}$ if $\eta(x)\geq \frac{1}{2}.$
	Recalling the definition of $f_c,$ we find that $f_P$ has the same sign of $f_c,$ that is Fisher consistency in statistics. In addition, for any $x\in X_c,$ $f(x)f_c(x)\leq 0.$ Therefore, $f(x)f_P(x)\leq 0,$ which implies that
	$0$ is between $f(x)$ and $f_P(x).$ This fact together with the convexity of $\Phi$ yields that
	$\Phi(f_P(x))\leq \Phi(0)\leq \Phi(f(x)).$
	
	Hence,
	\begin{eqnarray*}
		\mathcal{R}\big(\hbox{sgn}(f))-\mathcal{R}(f_c)&\leq & C_\tau^{-\frac{\tau}{\tau+2}}2^{1+\frac{(2q+1)(\tau+1)}
			{(q+1)(\tau+2)}}\Big(\frac{q+1}{q}\Big)^{\frac{\tau+1}{\tau+2}}
		\Big(\int_{X_c} (\Phi(f(x))-\Phi(f_P(x)))\ dP_X(x)\Big)^{\frac{\tau+1}{\tau+2}}\\
		&=& C_\tau^{-\frac{\tau}{\tau+2}}2^{1+\frac{(2q+1)(\tau+1)}{(q+1)(\tau+2)}}
		\Big(\frac{q+1}{q}\Big)^{\frac{\tau+1}{\tau+2}}(\mathcal{E}(f)-\mathcal{E}(f_P))^{\frac{\tau+1}{\tau+2}}.
	\end{eqnarray*}
	
	The proof of (\ref{thm.com5}) is complete.
\end{proof}
\section{Conclusion}
The LUM loss with $p=\infty$ reduces to the hinge loss, the comparison theorem of which has been well studied in \cite{Zhang2004}. In \cite{WangZou2018}, the comparison theorem was proved for generalized DWD with $p=q>0.$ In this paper, for the LUM loss with $p>0$, we obtain the best comparison theorem for any probability measure $P.$ However, when $p=0,$ things are essentially different. If we impose the Tsybakov noise condition on $P,$ the comparison theorem (\ref{thm.com2}) can be improved to (\ref{thm.com5}). Comparison theorem plays an important role for the error analysis of classification algorithms, which is also required for quantile regression \cite{Xiang12} and support vector regression \cite{XiangHuZhou11}.

\end{document}